\documentclass[runningheads]{llncs}
\usepackage{graphicx}

\usepackage{tikz}
\usepackage{comment}
\usepackage{amsmath,amssymb} 
\usepackage{color}

\usepackage[accsupp]{axessibility}  

\begin{document}
\pagestyle{headings}
\mainmatter
\def\ECCVSubNumber{6048}  

\title{Image Super-Resolution with Deep Dictionary} 


\titlerunning{Image Super-Resolution with Deep Dictionary}
%
\author{Shunta Maeda}
\authorrunning{S. Maeda}
%
\institute{Navier Inc.\\
\email{shunta@navier.co.jp}}
\maketitle

\begin{abstract}
\sloppy
Since the first success of Dong et al., the deep-learning-based approach has become dominant in the field of single-image super-resolution.
This replaces all the handcrafted image processing steps of traditional sparse-coding-based methods with a deep neural network.
In contrast to sparse-coding-based methods, which explicitly create high/low-resolution dictionaries, the dictionaries in deep-learning-based methods are implicitly acquired as a nonlinear combination of multiple convolutions.
One disadvantage of deep-learning-based methods is that their performance is degraded for images created differently from the training dataset (out-of-domain images).
We propose an end-to-end super-resolution network with a deep dictionary (SRDD), where a high-resolution dictionary is explicitly learned without sacrificing the advantages of deep learning.
Extensive experiments show that explicit learning of high-resolution dictionary makes the network more robust for out-of-domain test images while maintaining the performance of the in-domain test images.
Code is available at \textcolor{magenta}{\url{https://github.com/shuntama/srdd}}.

\end{abstract}

\section{Introduction}

\begin{figure}[t]
\centering
\includegraphics[height=2.58cm]{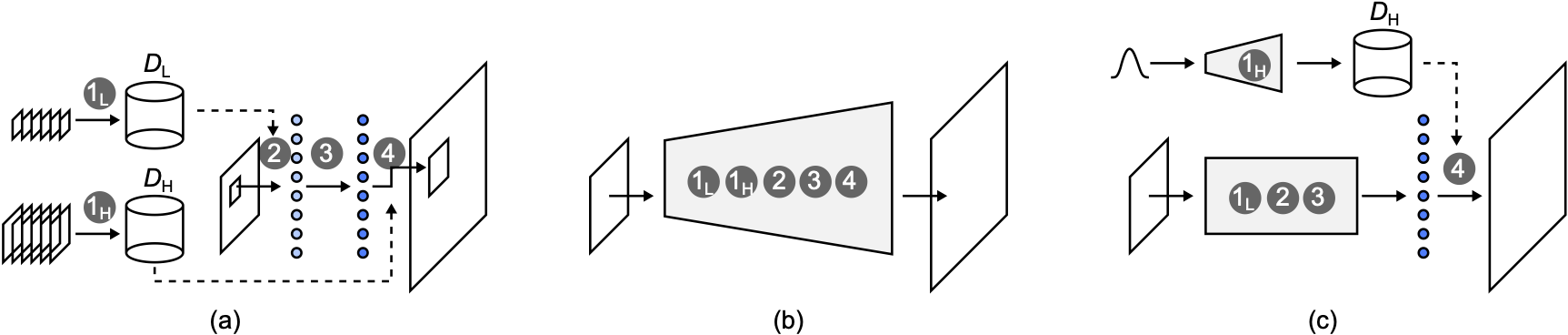}
\caption{Schematic illustrations of single image super-resolution with (a) sparse-coding-based approach, (b) conventional deep-learning-based approach, and (c) our approach. The numbers {\normalsize \textcircled{\small 1}}–{\normalsize \textcircled{\small 4}} indicate each step of the super-resolution process.}
\label{fig:compare}
\end{figure}

Single-image super-resolution (SISR) is a classical problem in the field of computer vision that predicts a high-resolution (HR) image from its low-resolution (LR) observation.
Because this is an ill-posed problem with multiple possible solutions, obtaining a rich prior based on a large number of data points is beneficial for better prediction.
Deep learning is quite effective for such problems.
The performance of SISR has been significantly improved by using convolutional neural networks (CNN), starting with the pioneering work of Dong et al. in 2014~\cite{dong2014learning}.
Before the dominance of deep-learning-based methods~\cite{ahn2018fast,dong2014learning,dong2016accelerating,kim2016accurate,kim2016deeply,lai2017deep,liang2021swinir,lim2017enhanced,tai2017image,tai2017memnet,zhang2018image,zhang2018residual} in this field, example-based methods~\cite{chang2004super,freeman2002example,glasner2009super,huang2015single,kim2010single,timofte2013anchored,timofte2014a+,yang2012coupled,yang2010image} were mainly used for learning priors.
Among them, sparse coding, which is a representative example-based method, has shown state-of-the-art performance~\cite{timofte2013anchored,timofte2014a+,yang2012coupled}.
SISR using sparse coding comprises the following steps, as illustrated in Fig.~\ref{fig:compare}(a): {\large \textcircled{\small $1_{\text{L}}$}} learn an LR dictionary $D_{\text{L}}$ with patches extracted from LR images, {\large \textcircled{\small $1_{\text{H}}$}} learn an HR dictionary $D_{\text{H}}$ with patches extracted from HR images, {\large \textcircled{\small 2}} represent patches densely cropped from an input image with $D_{\text{L}}$, {\large \textcircled{\small 3}} map $D_{\text{L}}$ representations to $D_{\text{H}}$ representations, {\large \textcircled{\small 4}} reconstruct HR patches using $D_{\text{H}}$, then aggregate the overlapped HR patches to produce a final output.

As depicted in Fig.~\ref{fig:compare}(b), Dong et al.~\cite{dong2014learning} replaced all the above handcrafted steps with a multilayered CNN in their proposed method SRCNN to take advantage of the powerful capability of deep learning.
Note that, in this method, $D_{\text{L}}$ and $D_{\text{H}}$ are implicitly acquired through network training.
Since the SRCNN, various methods have been proposed to improve performance, for example, by deepening the network with residual blocks and skip connections~\cite{kim2016deeply,lim2017enhanced,tai2017image,zhang2018residual}, applying attention mechanisms~\cite{dai2019second,mei2021image,niu2020single,zhang2018image}, and using a transformer~\cite{chen2021pre,liang2021swinir}.
However, most of these studies, including state-of-the-art ones, follow the same formality as SRCNN from a general perspective, where all the processes in the sparse-coding-based methods are replaced by a multilayered network.

One disadvantage of deep-learning-based methods is that their performance is degraded for images created differently from the training dataset~\cite{gu2019blind}.
Although there have been several approaches to address this issue, such as training networks for multiple degradations~\cite{soh2020meta,wang2021unsupervised,xu2020unified,zhang2018learning,zhou2019kernel} and making models agnostic to degradations with iterative optimizations~\cite{gu2019blind,shocher2018zero}, it is also important to make the network structure more robust.
We hypothesize that $D_{\text{H}}$ implicitly learned inside a multilayered network is fragile to subtle differences in input images from the training time.
This hypothesis leads us to the method we propose.

In this study, we propose an end-to-end super-resolution network with a deep dictionary (SRDD), where $D_{\text{H}}$ is explicitly learned through the network training (Fig.~\ref{fig:compare}(c)).
The main network predicts the coefficients of $D_{\text{H}}$ and the weighted sum of the elements (or atoms) of $D_{\text{H}}$ produces an HR output.
This approach is fundamentally different from the conventional deep-learning-based approach, where the network has upsampling layers inside it.
The upsampling process of the proposed method is efficient because the pre-generated $D_{\text{H}}$ can be used as a magnifier for inference.
In addition, the main network does not need to maintain the information of the processed image at the pixel level in HR space. Therefore, the network can concentrate only on predicting the coefficients of $D_{\text{H}}$.
For in-domain test images, our method shows performance that is not as good as latest ones, but close to the conventional baselines (eg., CARN).
For out-of-domain test images, our method shows superior performance compared to conventional deep-learning-based methods.

\section{Related Works}

\subsection{Sparse-coding-based SR}

Before the dominance of deep-learning-based methods in the field of SISR, example-based methods showed state-of-the-art performance.
The example-based methods exploit internal self-similarity~\cite{freedman2011image,glasner2009super,huang2015single,yang2013fast} and/or external datasets~\cite{chang2004super,freeman2002example,kim2010single,timofte2013anchored,timofte2014a+,yang2012coupled,yang2010image}.
The use of external datasets is especially important for obtaining rich prior.
In the sparse-coding-based methods~\cite{timofte2013anchored,timofte2014a+,yang2012coupled,yang2010image}, which are state-of-the-art example-based methods, high/low-resolution patch pairs are extracted from external datasets to create high/low-resolution dictionaries $D_{\text{H}}$/$D_{\text{L}}$.
The patches cropped from an input image are encoded with $D_{\text{L}}$ and then projected onto $D_{\text{H}}$ via iterative processing, producing the final output with appropriate patch aggregation.

\subsection{Deep-learning-based SR}

\noindent
\textbf{Deep CNN }
All the handcrafted steps in the traditional sparse-coding-based approach were replaced with an end-to-end CNN in a fully feed-forward manner.
Early methods, including SRCNN~\cite{dong2014learning,kim2016accurate,kim2016deeply}, adopted pre-upsampling in which LR input images are first upsampled for the SR process.
Because the pre-upsampling is computationally expensive, post-upsampling is generally used in recent models~\cite{ahn2018fast,liang2021swinir,lim2017enhanced,mei2021image,zhang2018image}.
In post-upsampling, a transposed convolution or pixelshuffle~\cite{shi2016real} is usually used to upsample the features for final output.
Although there are many proposals to improve network architectures~\cite{li2022ntire,zhang2020aim}, the protocol that directly outputs SR images with post-upsampling has been followed in most of those studies.
Few studies have focused on the improvement of the upsampling strategy.
Tough some recent works~\cite{chan2021glean,chen2022real,zhou2022towards} leveraged the pre-trained latent features as a dictionary to improve output fidelity with rich textures, they used standard upsampling strategies in their proposed networks.

\noindent
\textbf{Convolutional sparse coding }
Although methods following SRCNN have been common in recent years, several fundamentally different approaches have been proposed before and after the proposal of SRCNN.
Convolutional sparse coding~\cite{gu2015convolutional,osendorfer2014image,simon2019rethinking,wang2015deep} is one of such methods that work on the entire image differently from traditional patch-based sparse coding.
The advantage of convolutional sparse coding is that it avoids the boundary effect in patch-based sparse coding.
However, it conceptually follows patch-based sparse coding in that the overall SR process is divided into handcrafted steps.
Consequently, its performance lags behind that of end-to-end feed-forward CNN.

\noindent
\textbf{Robust SR }
The performance of deep-learning-based SR is significantly affected by the quality of the input image, especially the difference in conditions from the training dataset~\cite{gu2019blind}.
Several approaches have been proposed to make the network more robust against in-domain test images by training with multiple degradations~\cite{soh2020meta,wang2021unsupervised,xu2020unified,zhang2018learning,zhou2019kernel}.
For robustness against out-of-domain test images, some studies aim to make the network agnostic to degradations~\cite{gu2019blind,shocher2018zero}.
In these methods, agnostics acquisition is generally limited to specific degradations; therefore, it is important to make the network structure itself more robust.

\section{Method}

\begin{figure}[t]
\centering
\includegraphics[height=7.45cm]{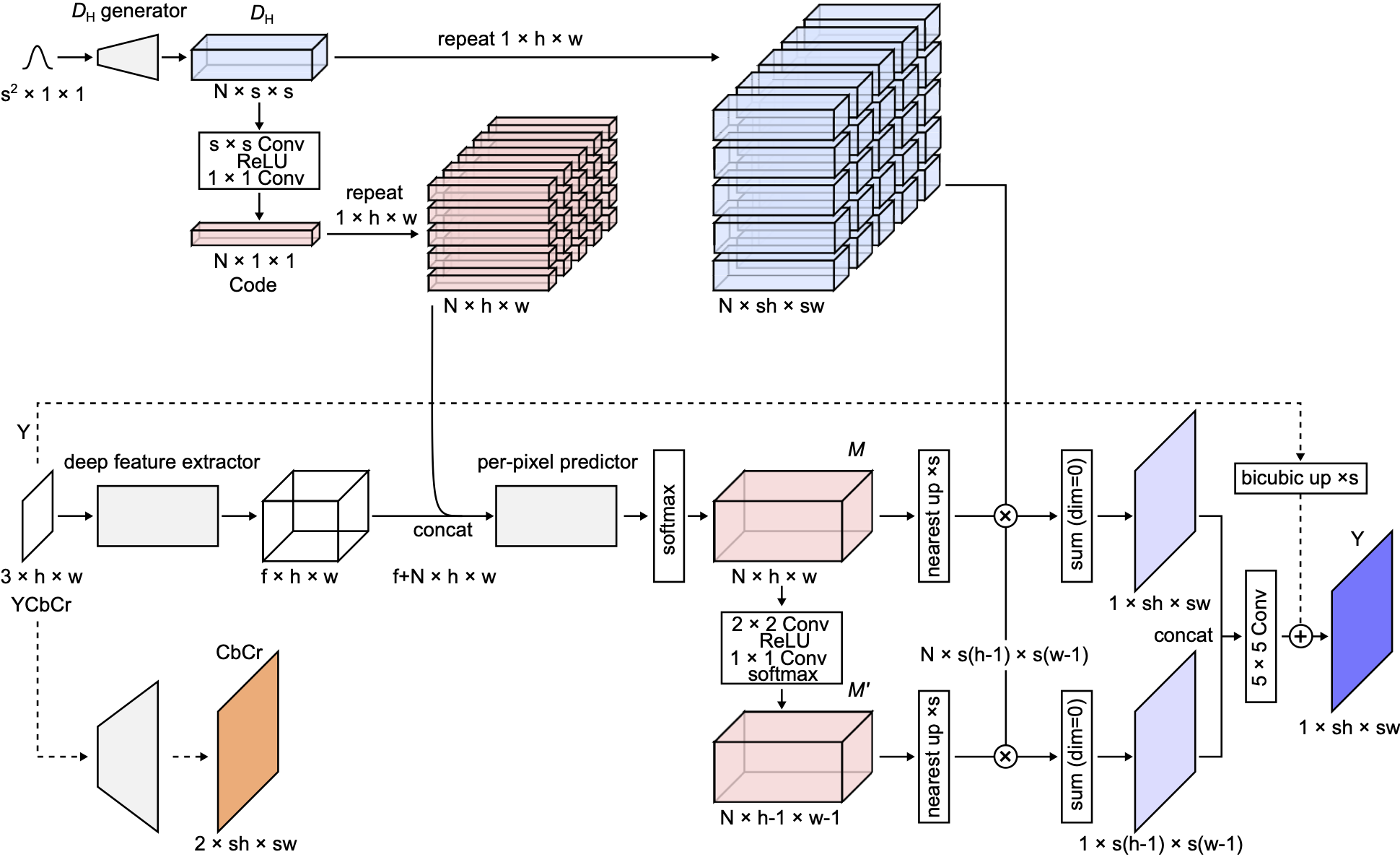}
\caption{The overall pipeline of the proposed method. A high-resolution dictionary $D_{\text{H}}$ is generated from random noise. An encoded code of $D_{\text{H}}$ is then concatenated with an extracted feature to be inputted to a per-pixel predictor. The predictor output is used to reconstruct the final output based on $D_{\text{H}}$.}
\label{fig:method}
\end{figure}

As depicted in Fig.~\ref{fig:compare}(c), the proposed method comprises three components: $D_{\text{H}}$ generation, per-pixel prediction, and reconstruction.
The $D_{\text{H}}$ generator generates an HR dictionary $D_{\text{H}}$ from random noise input.
The per-pixel predictor predicts the coefficients of $D_{\text{H}}$ for each pixel from an LR YCbCr input.
In the reconstruction part, the weighted sum of the elements (or atoms) of $D_{\text{H}}$ produces an HR Y-channel output as a residual to be added to a bicubically upsampled Y channel.
The remaining CbCr channels are upscaled with a shallow SR network.
We used ESPCN~\cite{shi2016real} as the shallow SR network in this work.
All of these components can be simultaneously optimized in an end-to-end manner; therefore, the same training procedure can be used as in conventional deep-learning-based SR methods.
We use $L_{1}$ loss function to optimize the network
\begin{align}
  L = \frac{1}{M} \sum_{i=1}^{M} ||I_i^{gt} - \Theta(I_i^{lr})||_{1},
\end{align}
where $I_i^{lr}$ and $I_i^{gt}$ are LR patch and its ground truth.
$M$ denotes the number of training image pairs.
$\Theta(\cdot)$ represents a function of the SRDD network.
Figure~\ref{fig:method} illustrates the proposed method in more detail.
We describe the design of each component based on Fig.~\ref{fig:method} in the following subsections.

\subsection{$D_{\text{H}}$ Generation}

From random noise $\delta^{s^2 \times 1 \times 1}$ ($\in\mathbb{R}^{s^2 \times 1 \times 1}$) with a standard normal distribution, the $D_{\text{H}}$ generator generates the HR dictionary $D_{\text{H}}^{N \times s \times s}$, where $s$ is an upscaling factor and $N$ is the number of elements (atoms) in the dictionary.
$D_{\text{H}}$ is then encoded by $s \times s$ convolution with groups $N$, followed by ReLU~\cite{nair2010rectified} and $1 \times 1$ convolution.
Each $N$ element of the resultant code $C_{\text{H}}^{N \times 1 \times 1}$ represents each $s \times s$ atom as a scalar value.
Although the $D_{\text{H}}$ can be trained using a fixed noise input, we found that introducing input randomness improves the stability of the training.
A pre-generated fixed dictionary and its code are used in the testing phase.
Note that only $D_{\text{H}}$ is generated since low-resolution dictionaries (encoding) can be naturally replaced by convolutional operations without excessive increases in computation.

\begin{figure}[t]
\centering
\includegraphics[height=3.8cm]{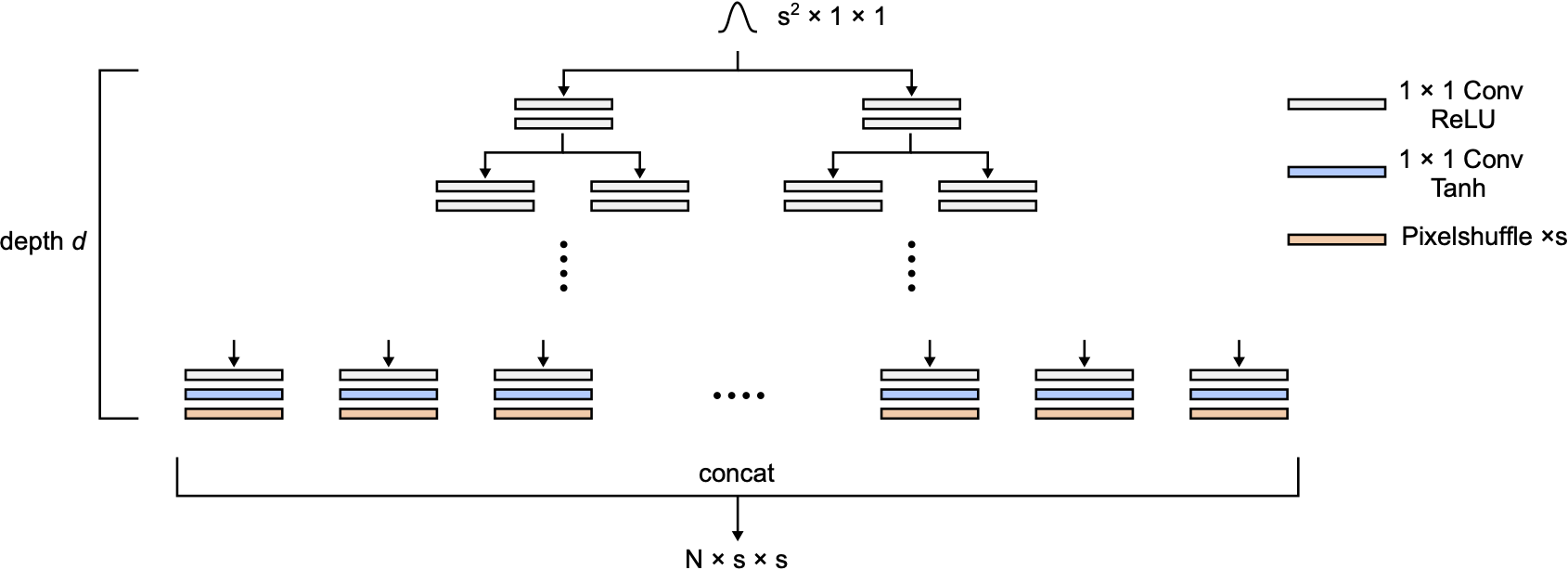}
\caption{A generator architecture of a high-resolution dictionary $D_{\text{H}}$. A tree-like network with depth $d$ generates $2^d$ atoms of size $1 \times s \times s$ from a random noise input, where $s$ is an upscaling factor.}
\label{fig:generator}
\end{figure}

\begin{figure}[t]
\centering
\includegraphics[height=1.21cm]{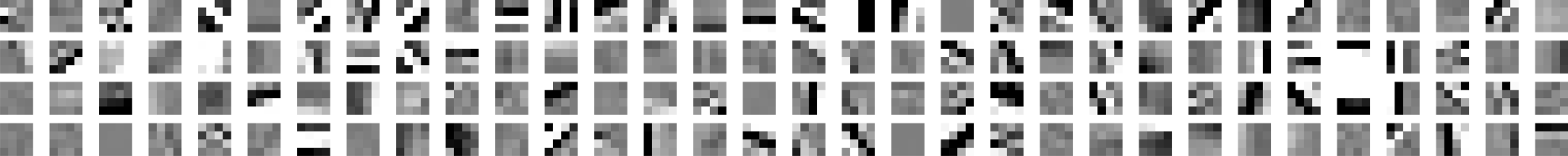}
\caption{Learned atoms of $\times 4$ SRDD with $N = 128$. The size of each atom is $1 \times 4 \times 4$. The data range is renormalized to $[0, 1]$ for visualization.}
\label{fig:atoms_x4}
\end{figure}

As illustrated in Fig.~\ref{fig:generator}, the $D_{\text{H}}$ generator has a tree-like structure, where the nodes consist of two $1 \times 1$ convolutional layers with ReLU activation.
The final layer has a Tanh activation followed by a pixelshuffling layer; therefore, the data range of the output atoms is $[-1, 1]$.
To produce $N$ atoms, depth $d$ of the generator is determined as
\begin{align}
  d = \log_2 N.
\end{align}
Figure~\ref{fig:atoms_x4} shows generated atoms with $s = 4$ and $N = 128$.
We observed that the contrast of the output atoms became stronger as training progressed, and they were almost fixed in the latter half of the training.

\subsection{Per-pixel Prediction}

We utilize UNet++~\cite{zhou2018unet++} as a deep feature extractor in Fig.~\ref{fig:method}.
We slightly modify the original UNet++ architecture: the depth is reduced from four to three, and a long skip connection is added.
The deep feature extractor outputs a tensor of size $f \times h \times w$ from the input YCbCr image, where $h$ and $w$ are the height and width of the image, respectively.  
Then the extracted feature is concatenated with the expanded code of $D_{\text{H}}$
\begin{align}
  C_{\text{H}}^{N \times h \times w} = R_{1 \times h \times w}(C_{\text{H}}^{N \times 1 \times 1}),
\end{align}
to be inputted to a per-pixel predictor, where $R_{a \times b \times c}(\cdot)$ denotes the $a \times b \times c$ repeat operations.
The per-pixel predictor consists of ten bottleneck residual blocks followed by a softmax function that predicts $N$ coefficients of $D_{\text{H}}$ for each input pixel.
Both the deep feature extractor and per-pixel predictor contain batch normalization layers~\cite{ioffe2015batch} before the ReLU activation.
The resultant prediction map $M^{N \times h \times w}$ is further convolved with a $2 \times 2$ convolution layer to produce a complementary prediction map $M'^{N \times (h-1) \times (w-1)}$.
A complementary prediction map is used to compensate for the patch boundaries when reconstructing the final output.
The detail of the compensation mechanism is described in the next subsection. 
Although we tried to replace softmax with ReLU to directly express sparsity, ReLU made the training unstable.
We also tried entmax~\cite{peters2019sparse}, but the performance was similar to that of softmax, so we decided to use softmax for simplicity.

\begin{figure}[t]
\centering
\includegraphics[height=2.8cm]{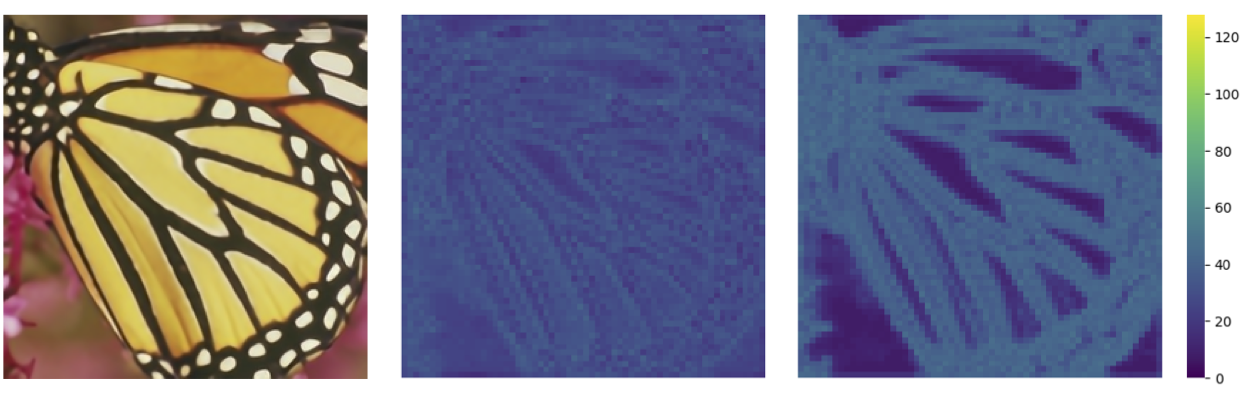}
\caption{Visualization of sparsity of a prediction map (center) and its complementary prediction map (right). The number of predicted coefficients larger than $1e-2$ is counted for each pixel. More atoms are assigned to the high-frequency parts and the low-frequency parts are relatively sparse.}
\label{fig:map}
\end{figure}

Figure~\ref{fig:map} visualizes the sparsity of the prediction map and its complementary prediction map.
The number of coefficients larger than $1e-2$ is counted for each pixel to visualize the sparsity.
The model with $N = 128$ is used.
More atoms are assigned to the high-frequency parts of the image, and the low-frequency parts are relatively sparse.
This feature is especially noticeable in the complementary prediction map.
In the high-frequency region, the output image is represented by linear combinations of more than tens of atoms for both maps.

\subsection{Reconstruction}

\begin{figure}[t]
\centering
\includegraphics[height=3.15cm]{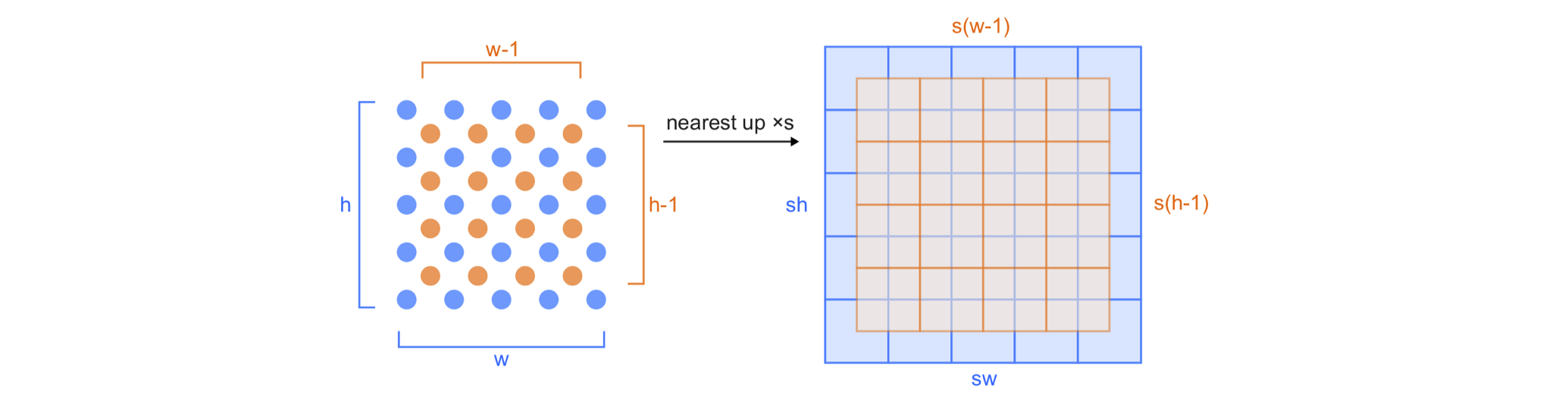}
\caption{Schematic illustration of a mechanism to compensate patch boundary with a complementary prediction map, where $s$ is a scaling factor. Left: Prediction map (blue) and its complementary prediction map (orange). Right: Upsampled prediction and complementary prediction maps with centering.}
\label{fig:compensate}
\end{figure}

The prediction map $M^{N \times h \times w}$ is upscaled to
$N \times sh \times sw$ by nearest-neighbor interpolation, and the element-wise multiplication of that upscaled prediction map $U_s(M^{N \times h \times w})$ with the expanded dictionary $R_{1 \times h \times w}(D_{\text{H}}^{N \times s \times s})$ produces $N \times sh \times sw$ tensor $T$ consists of weighted atoms.
The $U_s(\cdot)$ denotes $\times s$ nearest-neighbor upsampling.
Finally, tensor $T$ is summed over the first dimension, producing output $x$ as
\begin{eqnarray}
  x^{1 \times sh \times sw} &=& \sum_{k=0}^{N-1} T^{N \times sh \times sw}[k, :, :],\\
  T^{N \times sh \times sw} &=& U_s(M^{N \times h \times w}) \otimes R_{1 \times h \times w}(D_{\text{H}}^{N \times s \times s}).
\end{eqnarray}
The same sequence of operations is applied to the complementary prediction map to obtain the output $x'$ as follows:
\begin{eqnarray}
  x'^{1 \times s(h-1) \times s(w-1)} &=& \sum_{k=0}^{N-1} T'^{N \times s(h-1) \times s(w-1)}[k, :, :],\\
  T'^{N \times s(h-1) \times s(w-1)} &=& U_s(M'^{N \times (h-1) \times (w-1)}) \otimes R_{1 \times (h-1) \times (w-1)}(D_{\text{H}}^{N \times s \times s}).
\end{eqnarray}
Note that the same dictionary, $D_{\text{H}}$, is used to obtain $x$ and $x'$.
By centering $x$ and $x'$, as illustrated in Fig.~\ref{fig:compensate}, the imperfections at the patch boundaries can complement each other.
The final output residual is obtained by concatenating the overlapping parts of the centered $x$ and $x'$ and applying a $5 \times 5$ convolution.
For non-overlapping parts, $x$ is simply used as the final output.

\section{Experiments}

\subsection{Implementation Details}

We adopt a model with 128 atoms (SRDD-128) and a small model with 64 atoms (SRDD-64).
The number of filters of the models is adjusted according to the number of atoms.
Our network is trained by inputting $48 \times 48$ LR patches with a mini-batch size of 32.
Following previous studies~\cite{ahn2018fast,lim2017enhanced,zhang2018image}, random flipping and rotation augmentation is applied to each training sample.
We use Adam optimizer~\cite{kingma2014adam} with $\beta_1 = 0.9$, $\beta_2 = 0.999$, and $\epsilon = 10^{-8}$.
The learning rate of the network except for the $D_{\text{H}}$ generator is initialized as $2e-4$ and halved at [200k, 300k, 350k, 375k].
The total training iterations is 400k.
The learning rate of the $D_{\text{H}}$ generator is initialized as $5e-3$ and halved at [50k, 100k, 200k, 300k, 350k].
Parameters of the $D_{\text{H}}$ generator are frozen at 360k iteration.
In addition, to stabilize training of the $D_{\text{H}}$ generator, we randomly shuffle the order of output atoms for the first 1k iterations.
We use PyTorch to implement our model with an NVIDIA P6000 GPU.
Training takes about two/three days for SRDD-64/128, respectively.
More training details are provided in the supplementary material.

\subsection{Dataset and Evaluation}

\noindent
\textbf{Training dataset }
Following previous studies, we use 800 HR-LR image pairs of the DIV2K~\cite{timofte2017ntire} training dataset to train our models.
LR images are created from HR images by Matlab bicubic downsampling.
For validation, we use initial ten images from the DIV2K validation dataset.

\noindent
\textbf{Test dataset }
For testing, we evaluate the models on five standard benchmarks: Set5~\cite{bevilacqua2012low}, Set14~\cite{zeyde2010single}, BSD100~\cite{martin2001database}, Urban100~\cite{huang2015single}, and Manga109~\cite{matsui2017sketch}.
In addition to standard test images downsampled with Matlab bicubic function same as in training, we use test images that downsampled by OpenCV bicubic, bilinear, and area functions to evaluate the robustness of the models.
In addition, we evaluate the models on real-world ten historical photographs.

\noindent
\textbf{Evaluation }
We use common image quality metrics peak signal-to-noise ratio (PSNR) and structural similarity index (SSIM)~\cite{wang2004image} calculated on the Y channel (luminance channel) of YCbCr color space.
For evaluation of real-world test images, no-reference image quality metric NIQE~\cite{mittal2012making} is used since there are no ground-truth images.
Following previous studies, we ignore $s$ pixels from the border to calculate all the metrics, where $s$ is an SR scale factor.

\subsection{Ablation Study}

We conduct ablation experiments to examine the impact of individual elements in SRDD.
We report the results of SRDD-64 throughout this section.
The results of the ablation experiments on Set14 downsampled with Matlab bicubic function are summarized in Tab.~\ref{tab:ablations}.

\setlength{\tabcolsep}{10pt}
\begin{table}[t]
\scriptsize
\begin{center}
\caption{Results of ablation experiments on Set14 downsampled with Matlab bicubic function.}
\label{tab:ablations}
\begin{tabular}{lcc}
\hline
 & PSNR & SSIM\\
\hline
\hline
SRDD-64  & 28.54 & 0.7809\\
SRDD-64 $-$ batch norm. & 28.49 & 0.7792\\
SRDD-64 $-$ bottleneck blocks & 28.48 & 0.7790\\
SRDD-64 $-$ compensation & 28.51 & 0.7801\\
\hline
\end{tabular}
\end{center}
\end{table}

\begin{figure}[t]
\centering
\includegraphics[height=4.2cm]{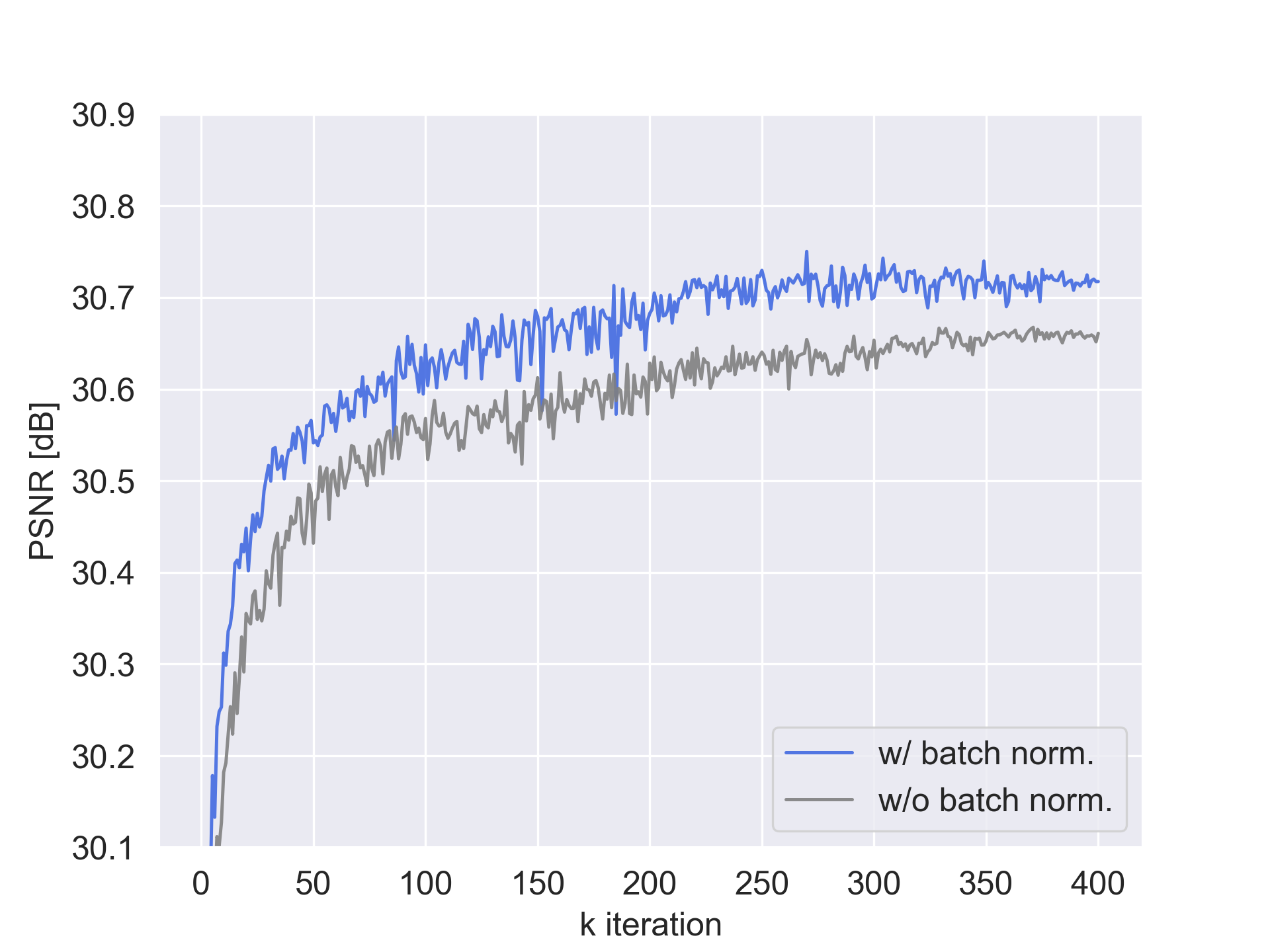}
\caption{Validation curves during the training of SRDD-64 with and without batch normalization layers.}
\label{fig:valid}
\end{figure}

\noindent
\textbf{Batch normalization }
We show the validation curves of SRDD-64 with and without batch normalization layers in Fig.~\ref{fig:valid}.
The performance of the proposed model is substantially improved by using batch normalization.
This result is in contrast to conventional deep-learning-based SR methods, where batch normalization generally degrades performance~\cite{lim2017enhanced}.
Unlike conventional methods where the network directly outputs the SR image, the prediction network in SRDD predicts the coefficients of $D_{\text{H}}$ for each pixel, which is rather similar to the semantic segmentation task.
In this sense, it is natural that batch normalization, which is effective for semantic segmentation~\cite{chen2017rethinking,li2018pyramid,zhao2017pyramid}, is also effective for the proposed model.

\noindent
\textbf{Bottleneck blocks }
We eliminate bottleneck blocks and $D_{\text{H}}$ code injection from the per-pixel predictor.
The prediction network becomes close to the plane UNet++ structure with this modification.
The performance drops as shown in Tab.~\ref{tab:ablations}, but still demonstrates a certain level of performance.

\noindent
\textbf{Compensation mechanism }
As shown in Tab.~\ref{tab:ablations}, removing the compensation mechanism from SRDD-64 degrades the performance.
However, the effect is marginal indicates that our model can produce adequate quality outputs without boundary compensation.
This result is in contrast to the sparse-coding-based methods, which generally require aggregation with overlapping patch sampling to reduce imperfection at the patch boundary.
Because the computational complexity of our compensation mechanism is very small compared to that of the entire model, we adopt it even if the effect is not so large.

\subsection{Results on In-domain Test Images}

We conduct experiments on five benchmark datasets, where the LR input images are created by Matlab bicubic downsampling same as in the DIV2K training dataset.
Because SRDD is quite shallow and fast compared to current state-of-the-art models, we compare SRDD to relatively shallow and fast models with roughly 50 layers or less.
Note that recent deep SR models usually have hundreds of convolutional layers~\cite{zhang2018image}.
We select ten models for comparison: SRCNN~\cite{dong2014learning}, FSRCNN~\cite{dong2016accelerating}, VDSR~\cite{kim2016accurate}, DRCN~\cite{kim2016deeply}, LapSRN~\cite{lai2017deep}, DRRN~\cite{tai2017image}, MemNet~\cite{tai2017memnet}, CARN~\cite{ahn2018fast}, IMDN~\cite{hui2019lightweight}, and LatticeNet~\cite{luo2020latticenet}.
We also compare our model with a representative sparse coding-based method A+~\cite{timofte2014a+}.
Results for the representative very deep models RCAN~\cite{zhang2018image}, NLSA~\cite{mei2021image}, and SwinIR~\cite{liang2021swinir} are also shown.

\setlength{\tabcolsep}{3.8pt}
\begin{table}[t]
\scriptsize
\begin{center}
\caption{Quantitative comparison for $\times4$ SR on benchmark datasets. Best and second best results are highlighted in \textcolor{red}{red} and \textcolor{blue}{blue}, respectively.}
\label{tab:psnr_ssim_x4_BI}
\begin{tabular}{l|c|c|c|c|c}
\hline
Method & Set5 &  Set14 &  BSD100 & Urban100 & Manga109  
\\
& PSNR / SSIM & PSNR / SSIM & PSNR / SSIM & PSNR / SSIM & PSNR / SSIM 
\\
\hline
\hline
Bicubic
& 28.42 / 0.8104
 & 26.00 / 0.7027
  & 25.96 / 0.6675
   & 23.14 / 0.6577
    & 24.89 / 0.7866
\\
A+~\cite{timofte2014a+}
& 30.28 / 0.8603
 & 27.32 / 0.7491
  & 26.82 / 0.7087
   & 24.32 / 0.7183
    & - / -
\\
SRCNN~\cite{dong2014learning}
& 30.48 / 0.8628
 & 27.50 / 0.7513
  & 26.90 / 0.7101
   & 24.52 / 0.7221
    & 27.58 / 0.8555
\\
FSRCNN~\cite{dong2016accelerating}
& 30.72 / 0.8660
 & 27.61 / 0.7550
  & 26.98 / 0.7150
   & 24.62 / 0.7280
    & 27.90 / 0.8610
\\
VDSR~\cite{kim2016accurate}
& 31.35 / 0.8830
 & 28.02 / 0.7680
  & 27.29 / 0.0726
   & 25.18 / 0.7540
    & 28.83 / 0.8870
\\
DRCN~\cite{kim2016deeply}
& 31.53 / 0.8854
 & 28.02 / 0.7670
  & 27.23 / 0.7233
   & 25.14 / 0.7510
    & - / -
\\
LapSRN~\cite{lai2017deep}
& 31.54 / 0.8850
 & 28.19 / 0.7720
  & 27.32 / 0.7270
   & 25.21 / 0.7560
    & 29.09 / 0.8900
\\
DRRN~\cite{tai2017image}
& 31.68 / 0.8888
 & 28.21 / 0.7720
  & 27.38 / 0.7284
   & 25.44 / 0.7638
    & - / -
\\
MemNet~\cite{tai2017memnet}
& 31.74 / 0.8893
 & 28.26 / 0.7723
  & 27.40 / 0.7281
   & 25.50 / 0.7630
    & 29.42 / 0.8942
\\
CARN~\cite{ahn2018fast}
& 32.13 / 0.8937
 & 28.60 / 0.7806
  & 27.58 / 0.7349
   & 26.07 / 0.7837
    & \textcolor{red}{30.47} / \textcolor{red}{0.9084}
\\
IMDN~\cite{hui2019lightweight}
& 32.21 / 0.8948
 & 28.58 / 0.7811
  & 27.56 / 0.7353
   & 26.04 / 0.7838
    & \textcolor{blue}{30.45} / \textcolor{blue}{0.9075}
\\
LatticeNet~\cite{luo2020latticenet}
& \textcolor{red}{32.30} / \textcolor{red}{0.8962}
 & \textcolor{red}{28.68} / \textcolor{blue}{0.7830}
  & \textcolor{red}{27.62} / \textcolor{blue}{0.7367}
   & \textcolor{red}{26.25} / \textcolor{blue}{0.7873}
    & - / -
\\
\hline
SRDD-64
& 32.05 / 0.8936
 & 28.54 / 0.7809
  & 27.54 / 0.7353
   & 25.89 / 0.7812
    & 30.16 / 0.9043
\\
SRDD-128
& \textcolor{blue}{32.25} / \textcolor{blue}{0.8958}
 & \textcolor{blue}{28.65} / \textcolor{red}{0.7838}
  & \textcolor{blue}{27.61} / \textcolor{red}{0.7378}
   & \textcolor{blue}{26.10} / \textcolor{red}{0.7877}
    & 30.44 / \textcolor{red}{0.9084}
\\
\hline
RCAN~\cite{zhang2018image}
& (32.63/0.9002)
 & (28.87/0.7889)
  & (27.77/0.7436)
   & (26.82/0.8087)
    & (31.22/0.9173)
\\
NLSA~\cite{mei2021image}
& (32.59/0.9000)
 & (28.87/0.7891)
  & (27.78/0.7444)
   & (26.96/0.8109)
    & (31.27/0.9184)
\\
SwinIR~\cite{liang2021swinir}
& (32.72/0.9021)
 & (28.94/0.7914)
  & (27.83/0.7459)
   & (27.07/0.8164)
    & (31.67/0.9226)
\\
\hline
\end{tabular}
\end{center}
\end{table}

\setlength{\tabcolsep}{12pt}
\begin{table}[t]
\scriptsize
\begin{center}
\caption{Execution time of representative models on an Nvidia P4000 GPU for $\times 4$ SR with input size $256\times256$.}
\label{tab:speed_x4}
\begin{tabular}{lcc}
\hline
 & Running time [s]\\
\hline
\hline
SRCNN~\cite{dong2014learning} & 0.0669\\
FSRCNN~\cite{dong2016accelerating} & 0.0036\\
VDSR~\cite{kim2016accurate} & 0.2636\\
LapSRN~\cite{lai2017deep} & 0.1853\\
CARN~\cite{ahn2018fast} & 0.0723\\
IMDN~\cite{hui2019lightweight} & 0.0351\\
\hline
SRDD-64     & 0.0842\\
SRDD-128    & 0.2196\\
\hline
RCAN~\cite{zhang2018image} & 1.5653\\
NLSA~\cite{mei2021image} & 1.7139\\
SwinIR~\cite{liang2021swinir} & 2.1106\\
\hline
\end{tabular}
\end{center}
\end{table}

\begin{figure}[t]
\scriptsize
\centering
    \setlength{\tabcolsep}{1pt}
	\begin{tabular}{c}
		\includegraphics[width=0.16 \textwidth]{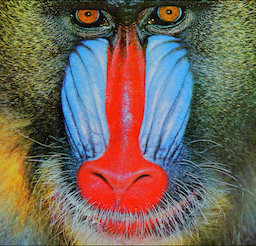}
		\\
		Set14: \textit{baboon}
		\\
	\end{tabular}
	\begin{tabular}{cccccc}
		\includegraphics[width=0.1 \textwidth]{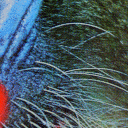} &
		\includegraphics[width=0.1 \textwidth]{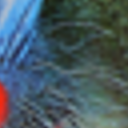} &
		\includegraphics[width=0.1 \textwidth]{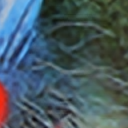} &
		\includegraphics[width=0.1 \textwidth]{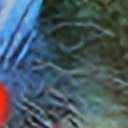} &
		\includegraphics[width=0.1 \textwidth]{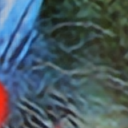} 
		\\
		HR &
		Bicubic &
		A+ &
		SRCNN &
		FSRCNN 
		\\
		\includegraphics[width=0.1 \textwidth]{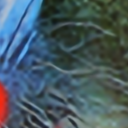} &
		\includegraphics[width=0.1 \textwidth]{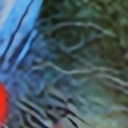} &
		\includegraphics[width=0.1 \textwidth]{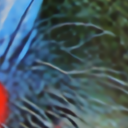} &
		\includegraphics[width=0.1 \textwidth]{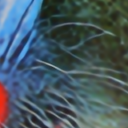} &
		\includegraphics[width=0.1 \textwidth]{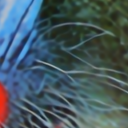}  
		\\
		VDSR &
		LapSRN &
		CARN &
		SRDD-64 &
		SRDD-128
		\\
	\end{tabular}
	
	\begin{tabular}{c}
		\includegraphics[width=0.16 \textwidth]{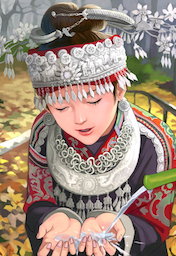}
		\\
		Set14: \textit{comic}
		\\
	\end{tabular}
	\begin{tabular}{cccccc}
		\includegraphics[width=0.1 \textwidth]{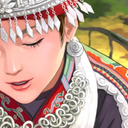} &
		\includegraphics[width=0.1 \textwidth]{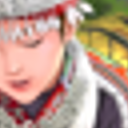} &
		\includegraphics[width=0.1 \textwidth]{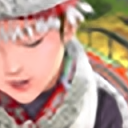} &
		\includegraphics[width=0.1 \textwidth]{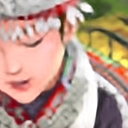} &
		\includegraphics[width=0.1 \textwidth]{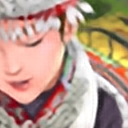} 
		\\
		HR &
		Bicubic &
		A+ &
		SRCNN &
		FSRCNN 
		\\
		\includegraphics[width=0.1 \textwidth]{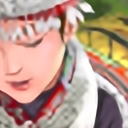} &
		\includegraphics[width=0.1 \textwidth]{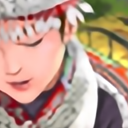} &
		\includegraphics[width=0.1 \textwidth]{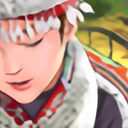} &
		\includegraphics[width=0.1 \textwidth]{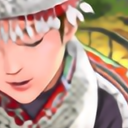} &
		\includegraphics[width=0.1 \textwidth]{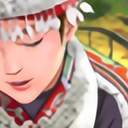}  
		\\ 
		VDSR &
		LapSRN &
		CARN &
		SRDD-64 &
		SRDD-128
		\\
	\end{tabular}
	
	\begin{tabular}{c}
		\includegraphics[width=0.16 \textwidth]{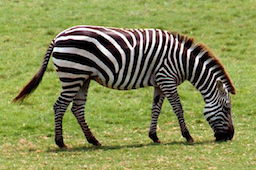}
		\\
		Set14: \textit{zebra}
		\\
	\end{tabular}
	\begin{tabular}{cccccc}
		\includegraphics[width=0.1 \textwidth]{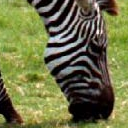} &
		\includegraphics[width=0.1 \textwidth]{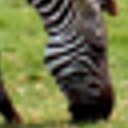} &
		\includegraphics[width=0.1 \textwidth]{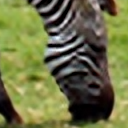} &
		\includegraphics[width=0.1 \textwidth]{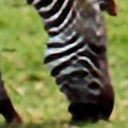} &
		\includegraphics[width=0.1 \textwidth]{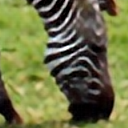} 
		\\
		HR &
		Bicubic &
		A+ &
		SRCNN &
		FSRCNN 
		\\
		\includegraphics[width=0.1 \textwidth]{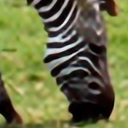} &
		\includegraphics[width=0.1 \textwidth]{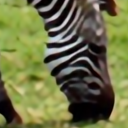} &
		\includegraphics[width=0.1 \textwidth]{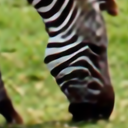} &
		\includegraphics[width=0.1 \textwidth]{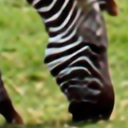} &
		\includegraphics[width=0.1 \textwidth]{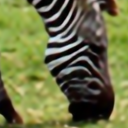}  
		\\ 
		VDSR &
		LapSRN &
		CARN &
		SRDD-64 &
		SRDD-128
		\\
	\end{tabular}
	
	\begin{tabular}{c}
		\includegraphics[width=0.16 \textwidth]{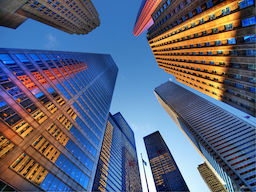}
		\\
		Urban100: \textit{012}
		\\
	\end{tabular}
	\begin{tabular}{cccccc}
		\includegraphics[width=0.1 \textwidth]{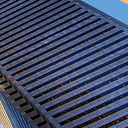} &
		\includegraphics[width=0.1 \textwidth]{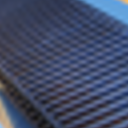} &
		\includegraphics[width=0.1 \textwidth]{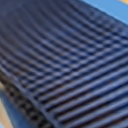} &
		\includegraphics[width=0.1 \textwidth]{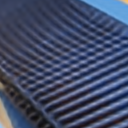} &
		\includegraphics[width=0.1 \textwidth]{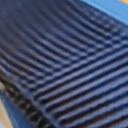} 
		\\
		HR &
		Bicubic &
		A+ &
		SRCNN &
		FSRCNN 
		\\
		\includegraphics[width=0.1 \textwidth]{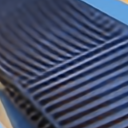} &
		\includegraphics[width=0.1 \textwidth]{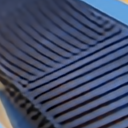} &
		\includegraphics[width=0.1 \textwidth]{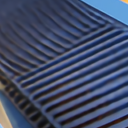} &
		\includegraphics[width=0.1 \textwidth]{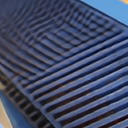} &
		\includegraphics[width=0.1 \textwidth]{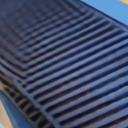}  
		\\ 
		VDSR &
		LapSRN &
		CARN &
		SRDD-64 &
		SRDD-128
		\\
	\end{tabular}
	
	\begin{tabular}{c}
		\includegraphics[width=0.16 \textwidth]{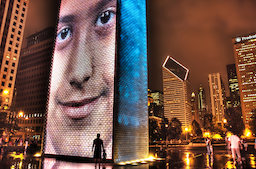}
		\\
		Urban100: \textit{076}
		\\
	\end{tabular}
	\begin{tabular}{cccccc}
		\includegraphics[width=0.1 \textwidth]{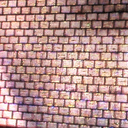} &
		\includegraphics[width=0.1 \textwidth]{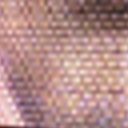} &
		\includegraphics[width=0.1 \textwidth]{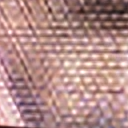} &
		\includegraphics[width=0.1 \textwidth]{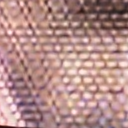} &
		\includegraphics[width=0.1 \textwidth]{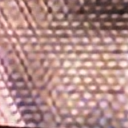} 
		\\
		HR &
		Bicubic &
		A+ &
		SRCNN &
		FSRCNN 
		\\
		\includegraphics[width=0.1 \textwidth]{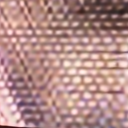} &
		\includegraphics[width=0.1 \textwidth]{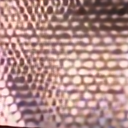} &
		\includegraphics[width=0.1 \textwidth]{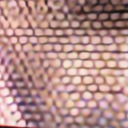} &
		\includegraphics[width=0.1 \textwidth]{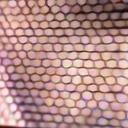} &
		\includegraphics[width=0.1 \textwidth]{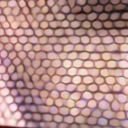}  
		\\ 
		VDSR &
		LapSRN &
		CARN &
		SRDD-64 &
		SRDD-128
		\\
	\end{tabular}
\caption{Visual comparison for $\times4$ SR on Set14 and Urban100 dataset. Zoom in for a better view.}
\label{fig:result_4x_BI}
\end{figure}

The quantitative results for $\times4$ SR on benchmark datasets are shown in Tab.~\ref{tab:psnr_ssim_x4_BI}.
SRDD-64 and SRDD-128 show comparable performances to CARN/IMDN and LatticeNet, respectively.
As shown in Tab.~\ref{tab:speed_x4}, the inference speed of SRDD-64 is also comparable to that of CARN, but slower than IMDN.
These results indicate that the overall performance of our method on in-domain test images is close to that of conventional baselines (not as good as state-of-the-art models).
The running time of representative deep models are also shown for comparison.
They are about 20 times slower than CARN and SRDD-64.
The visual results are provided in Fig.~\ref{fig:result_4x_BI}.

\clearpage

\subsection{Results on Out-of-domain Test Images}

\noindent
\textbf{Synthetic test images }
We conduct experiments on Set14, where the LR input images are created differently from training time.
We use bicubic, bilinear, and area downsampling with OpenCV resize functions.
The difference between Matlab and OpenCV resize functions mainly comes from the anti-aliasing option.
The anti-aliasing is default enabled/unenabled in Matlab/OpenCV, respectively.
We mainly evaluate CARN and SRDD-64 because these models have comparable performance on in-domain test images.
The state-of-the-art lightweight model IMDN~\cite{hui2019lightweight} and the representative blind SR model IKC~\cite{gu2019blind} are also evaluated for comparison.
The results are shown in Tab.~\ref{tab:psnr_ssim_x4_opencv}.
SRDD-64 outperformed these models by a large margin for the three different resize functions.
This result implies that our method is more robust for the out-of-domain images than conventional deep-learning-based methods.
The visual comparison on a test image downsampled with OpenCV bicubic function is shown in Fig.~\ref{fig:opencv_bicubic}.
CARN overly emphasizes high-frequency components of the image, while SRDD-64 outputs a more natural result.

\setlength{\tabcolsep}{5pt}
\begin{table}[t]
\scriptsize
\begin{center}
\caption{Quantitative results of $\times4$ SR on Set14 downsampled with three different OpenCV resize functions. Note that the models are trained with Matlab bicubic downsampling.}
\label{tab:psnr_ssim_x4_opencv}
\begin{tabular}{lccc}
\hline
& Bicubic & Bilinear & Area\\
& PSNR / SSIM & PSNR / SSIM & PSNR / SSIM\\
\hline
\hline
CARN~\cite{ahn2018fast} & 21.17 / 0.6310 & 22.76 / 0.6805 & 26.74 / 0.7604\\
IMDN~\cite{hui2019lightweight} & 20.99 / 0.6239 & 22.54 / 0.6741 & 26.60 / 0.7589\\
IKC~\cite{gu2019blind} & 20.10 / 0.6031 & 21.71 / 0.6558 & 26.40 / 0.7554\\
SRDD-64 & 21.52 / 0.6384 & 23.13 / 0.6871 & 27.05 / 0.7630\\
\hline
\end{tabular}
\end{center}
\end{table}

\begin{figure}[t]
\scriptsize
\centering
    \setlength{\tabcolsep}{1pt}
	\begin{tabular}{c}
		\includegraphics[height=0.16 \textwidth]{img_baboon_x4_GT_resize.png}
		\\
		\\
	\end{tabular}
	\begin{tabular}{ccc}
		\includegraphics[height=0.16 \textwidth]{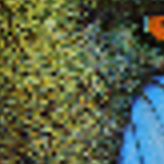} &
		\includegraphics[height=0.16 \textwidth]{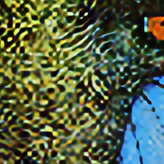} &
		\includegraphics[height=0.16 \textwidth]{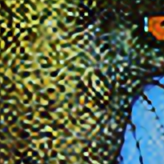}
		\\
		Bicubic &
		CARN &
		SRDD-64 
		\\
	\end{tabular}
\caption{Visual comparison for $\times4$ SR on Set14 \textit{baboon} downsampled with OpenCV bicubic function.}
\label{fig:opencv_bicubic}
\end{figure}

\noindent
\textbf{Real-world test images }
We conduct experiments on widely used ten historical images to see the robustness of the models on unknown degradations.
Because there is no ground-truth image, we adopt a no-reference image quality metric NIQE for evaluation.
Table~\ref{tab:niqe_x4_real} shows average NIQE for representative methods.
As seen in the previous subsection, our SRDD-64 shows comparable performance to CARN if compared with in-domain test images.
However, on the realistic datasets with the NIQE metric, SRDD-64 clearly outperforms CARN and is close to EDSR.
Interestingly, unlike the results on the in-domain test images, the performance of SRDD-64 is better than that of SRDD-128 for realistic degradations.
This is probably because representing an HR image with a small number of atoms makes the atoms more versatile.
The visual results are provided in Fig.~\ref{fig:result_4x_real}.

\setlength{\tabcolsep}{5pt}
\begin{table}[t]
\scriptsize
\begin{center}
\caption{Results of no-reference image quality metric NIQE on real-world historical images. Note that the models are trained with Matlab bicubic downsampling.}
\label{tab:niqe_x4_real}
\begin{tabular}{lc}
\hline
& NIQE (lower is better)\\
\hline
\hline
Bicubic & 7.342\\
A+~\cite{timofte2014a+} & 6.503\\
SRCNN~\cite{dong2014learning} & 6.267\\
FSRCNN~\cite{dong2016accelerating} & 6.130\\
VDSR~\cite{kim2016accurate} & 6.038\\
LapSRN~\cite{lai2017deep} & 6.234\\
CARN~\cite{ahn2018fast}	& 5.921\\
EDSR~\cite{lim2017enhanced}	& \textcolor{red}{5.852}\\
\hline
SRDD-64		& \textcolor{blue}{5.877}\\
SRDD-128	& 5.896\\
\hline
\end{tabular}
\end{center}
\end{table}

\begin{figure}[t]
\scriptsize
\centering
    \setlength{\tabcolsep}{1pt}
	\begin{tabular}{c}
		\includegraphics[width=0.16 \textwidth]{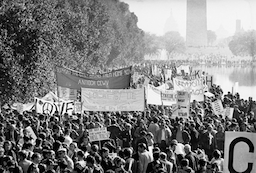}
		\\
		Historical: \textit{004}
		\\
	\end{tabular}
	\begin{tabular}{cccccc}
		\includegraphics[width=0.1 \textwidth]{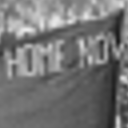} &
		\includegraphics[width=0.1 \textwidth]{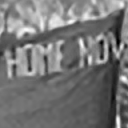} &
		\includegraphics[width=0.1 \textwidth]{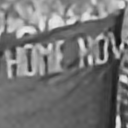} &
		\includegraphics[width=0.1 \textwidth]{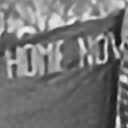} &
		\includegraphics[width=0.1 \textwidth]{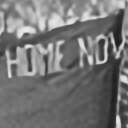} &
		\\
		Bicubic &
		A+ &
		SRCNN &
		FSRCNN &
		VDSR
		\\
		\includegraphics[width=0.1 \textwidth]{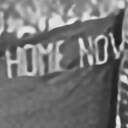} &
		\includegraphics[width=0.1 \textwidth]{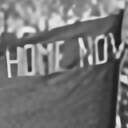} &
		\includegraphics[width=0.1 \textwidth]{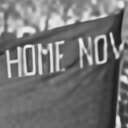} &
		\includegraphics[width=0.1 \textwidth]{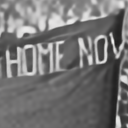} &
		\includegraphics[width=0.1 \textwidth]{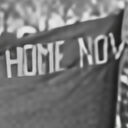}  
		\\
		LapSRN &
		CARN &
		EDSR &
		SRDD-64 &
		SRDD-128
		\\
	\end{tabular}
	
	\begin{tabular}{c}
		\includegraphics[width=0.16 \textwidth]{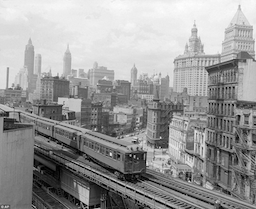}
		\\
		Historical: \textit{007}
		\\
	\end{tabular}
	\begin{tabular}{cccccc}
		\includegraphics[width=0.1 \textwidth]{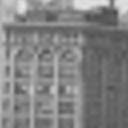} &
		\includegraphics[width=0.1 \textwidth]{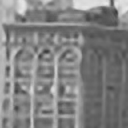} &
		\includegraphics[width=0.1 \textwidth]{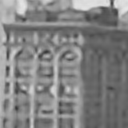} &
		\includegraphics[width=0.1 \textwidth]{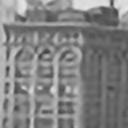} &
		\includegraphics[width=0.1 \textwidth]{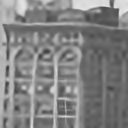} &
		\\
		Bicubic &
		A+ &
		SRCNN &
		FSRCNN &
		VDSR
		\\
		\includegraphics[width=0.1 \textwidth]{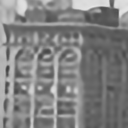} &
		\includegraphics[width=0.1 \textwidth]{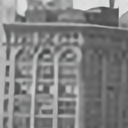} &
		\includegraphics[width=0.1 \textwidth]{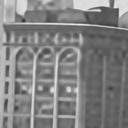} &
		\includegraphics[width=0.1 \textwidth]{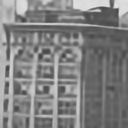} &
		\includegraphics[width=0.1 \textwidth]{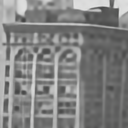}  
		\\
		LapSRN &
		CARN &
		EDSR &
		SRDD-64 &
		SRDD-128
		\\
	\end{tabular}
\caption{Visual comparison for $\times4$ SR on real-world historical images. Zoom in for a better view.}
\label{fig:result_4x_real}
\end{figure}

\subsection{Experiments of $\times$8 SR}

\begin{figure}[t]
\centering
\includegraphics[height=3.4cm]{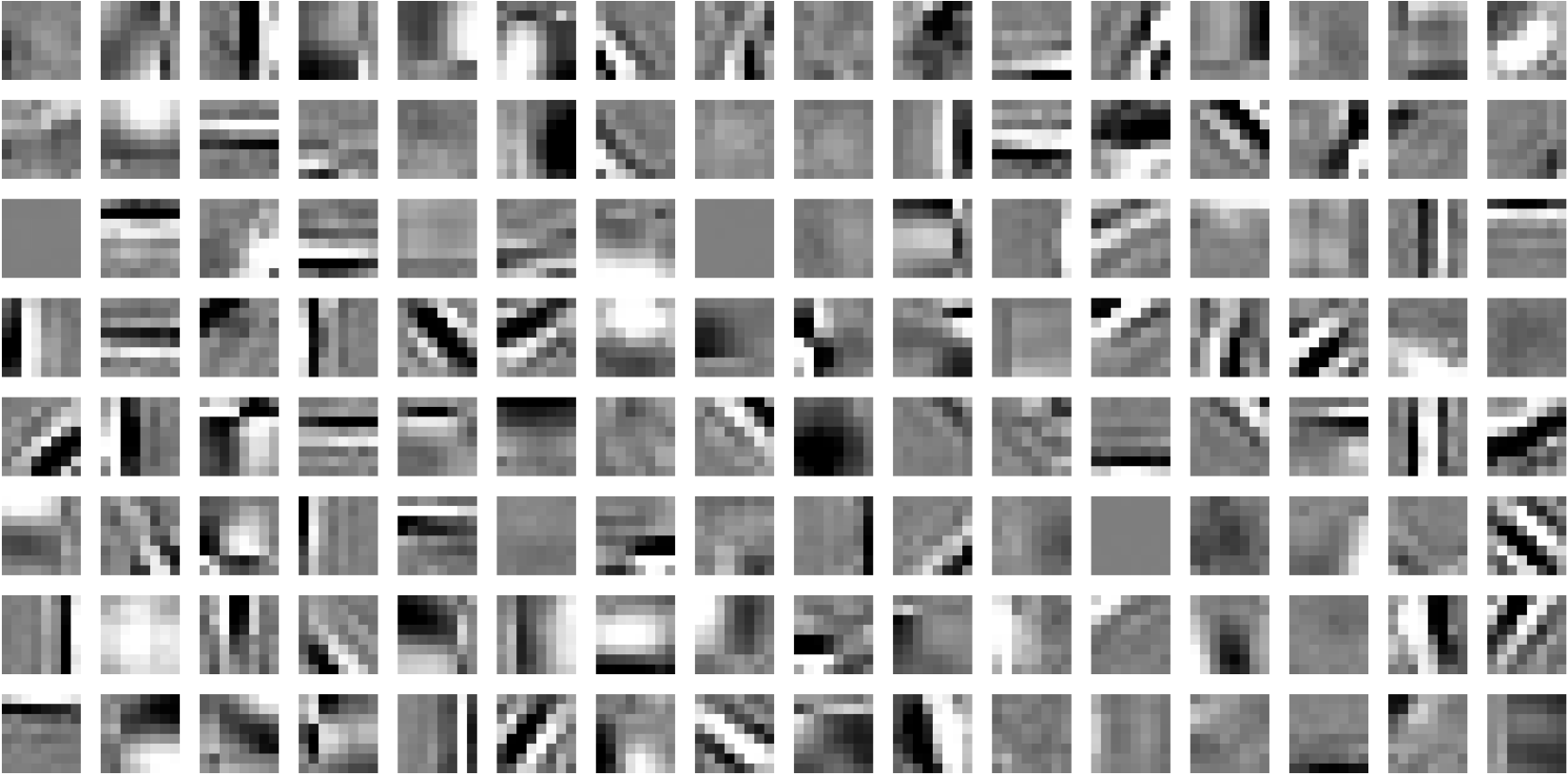}
\caption{Learned atoms of $\times 8$ SRDD with $N = 128$. The size of each atom is $1 \times 8 \times 8$. The data range is renormalized to $[0, 1]$ for visualization.}
\label{fig:atoms_x8}
\end{figure}

To see if our method would work at different scaling factors, we also experiment with the $\times 8$ SR case.
We use DIV2K dataset for training and validation.
The test images are prepared with the same downsampling function (i.e. Matlab bicubic function) as the training dataset.
Figure~\ref{fig:atoms_x8} shows generated atoms of SRDD with $s = 8$ and $N = 128$.
The structure of atoms with $s = 8$ is finer than that with $s = 4$, while the coarse structures of both cases are similar.
The quantitative results on five benchmark datasets are shown in Tab.~\ref{tab:psnr_ssim_x8_BI}.
SRDD performs better than the representative shallow models though its performance does not reach representative deep model EDSR.

\setlength{\tabcolsep}{4pt}
\begin{table}[t]
\scriptsize
\begin{center}
\caption{Quantitative comparison for $\times8$ SR on benchmark datasets. Best and second best results are highlighted in \textcolor{red}{red} and \textcolor{blue}{blue}, respectively.}
\label{tab:psnr_ssim_x8_BI}
\begin{tabular}{l|c|c|c|c|c}
\hline
Method & Set5 &  Set14 &  BSD100 & Urban100 & Manga109  
\\
& PSNR / SSIM & PSNR / SSIM & PSNR / SSIM & PSNR / SSIM & PSNR / SSIM 
\\
\hline
\hline
Bicubic
& 24.40 / 0.6580
 & 23.10 / 0.5660
  & 23.67 / 0.5480
   & 20.74 / 0.5160
    & 21.47 / 0.6500
\\
SRCNN~\cite{dong2014learning}
& 25.33 / 0.6900
 & 23.76 / 0.5910
  & 24.13 / 0.5660
   & 21.29 / 0.5440
    & 22.46 / 0.6950
\\
FSRCNN~\cite{dong2016accelerating}
& 20.13 / 0.5520
 & 19.75 / 0.4820
  & 24.21 / 0.5680
   & 21.32 / 0.5380
    & 22.39 / 0.6730
\\
VDSR~\cite{kim2016accurate}
& 25.93 / 0.7240
 & 24.26 / 0.6140
  & 24.49 / 0.5830
   & 21.70 / 0.5710
    & 23.16 / 0.7250
\\
LapSRN~\cite{lai2017deep}
& 26.15 / 0.7380
 & 24.35 / 0.6200
  & 24.54 / 0.5860
   & 21.81 / 0.5810
    & 23.39 / 0.7350
\\
MemNet~\cite{tai2017memnet}
& 26.16 / 0.7414
 & 24.38 / 0.6199
  & 24.58 / 0.5842
   & 21.89 / 0.5825
    & 23.56 / 0.7387
\\
EDSR~\cite{lim2017enhanced}
& \textcolor{red}{26.96} / \textcolor{red}{0.7762}
 & \textcolor{red}{24.91} / \textcolor{red}{0.6420}
  & \textcolor{red}{24.81} / \textcolor{red}{0.5985}
   & \textcolor{red}{22.51} / \textcolor{red}{0.6221}
    & \textcolor{red}{24.69} / \textcolor{red}{0.7841}
\\
\hline
SRDD-64
& 26.66 / 0.7652
 & 24.75 / 0.6345
  & 24.71 / 0.5926
   & 22.20 / 0.6034
    & 24.14 / 0.7621
\\
SRDD-128
& \textcolor{blue}{26.76} / \textcolor{blue}{0.7677}
 & \textcolor{blue}{24.79} / \textcolor{blue}{0.6369}
  & \textcolor{blue}{24.75} / \textcolor{blue}{0.5947}
   & \textcolor{blue}{22.25} / \textcolor{blue}{0.6073}
    & \textcolor{blue}{24.25} / \textcolor{blue}{0.7672}
\\
\hline
\end{tabular}
\end{center}
\end{table}

\section{Conclusions}
\sloppy
We propose an end-to-end super-resolution network with a deep dictionary (SRDD).
An explicitly learned high-resolution dictionary ($D_{\text{H}}$) is used to upscale the input image as in the sparse-coding-based methods, while the entire network, including the $D_{\text{H}}$ generator, is simultaneously optimized to take full advantage of deep learning.
For in-domain test images (images created by the same procedure as the training dataset), the proposed SRDD shows performance that is not as good as latest ones, but close to the conventional baselines (eg., CARN).
For out-of-domain test images, SRDD outperforms conventional deep-learning-based methods, demonstrating the robustness of our model.

The proposed method is not limited to super-resolution tasks but is potentially applicable to other tasks that require high-resolution output, such as high-resolution image generation.
Hence, the proposed method is expected to have a broad impact on various tasks.
Future works will be focused on the application of the proposed method to other vision tasks.
In addition, we believe that our method still has much room for improvement compared to the conventional deep-learning-based approach.

~

\noindent
\textbf{Acknowledgements }
I thank Uday Bondi for helpful comments on the manuscript.

\clearpage
%
%
\bibliographystyle{splncs04}
\bibliography{ms.bib}
\end{document}